\pdfoutput=1
\documentclass[11pt,a4paper]{article}
\usepackage[hyperref]{acl2021}
\usepackage{times}
\usepackage{latexsym}

\usepackage{graphicx}
\usepackage{microtype}
\usepackage{amsmath}
\usepackage{multirow}
\aclfinalcopy 
\def\aclpaperid{3285} 

\title{Modeling Language Usage and Listener Engagement in Podcasts}

\author{Sravana Reddy \\
  Spotify \\
  \texttt{sravana@spotify.com} \\\And
  Mariya Lazarova \\
  Spotify \\
  \texttt{mimiglazarova@gmail.com} \\\AND
  Yongze Yu \\
  Spotify \\
  \texttt{yongzey@spotify.com} \\\And
  Rosie Jones \\
  Spotify \\
  \texttt{rjones@spotify.com} 
  }

\date{}

\begin{document}
\maketitle

\begin{abstract}
 While there is an abundance of popular writing targeted to podcast creators on how to speak in ways that engage their listeners, there has been little data-driven analysis of podcasts that relates linguistic style with listener engagement.
In this paper, we investigate how various factors -- vocabulary diversity, distinctiveness, emotion, and syntax, among others -- correlate with engagement, based on analysis of the creators' written descriptions and transcripts of the audio. We build models with different textual representations, and show that the identified  features are highly predictive of engagement. Our analysis tests popular wisdom about stylistic elements in high-engagement podcasts, corroborating some aspects, and adding new perspectives on others. 
\end{abstract}

\section{Introduction}
What makes a particular podcast broadly engaging?  As a media form, podcasting is new enough that such questions are only beginning to be understood \cite{podcastperspectives}.
Websites exist with advice on podcast production, including language-related tips such as reducing filler words and disfluencies, or incorporating emotion, but there has been little quantitative research into how aspects of language usage contribute to listener engagement.

This paper investigates the linguistic factors that correlate with engagement, leveraging the written descriptions of the parent show and episode as well as the transcript of the audio. Our metric of engagement is {\bf stream rate}, which we define as the proportion of first-time listeners -- of those who have begun streaming the episode -- who listen for at least five minutes. Notably, stream rate is different from the metric of popularity as given by the raw number of streams; the latter is inevitably influenced by factors unrelated to the content, such as the host or publisher reputation, publicity, exposure in recommendations and search engines, and time of publication, whereas a listener's decision to  continue listening for as long as five minutes is likely to be influenced by the content.

We perform a series of descriptive tests to examine differences in language usage between high and low engagement podcasts, and build predictive models.
 Our tests show that while much of the conventional wisdom on engaging podcasting style (such as to use positive language) bears out in the data, other assumptions (such as to speak slowly) are contradicted and deserve a closer look. 
 We find that stylistic features tend to be more  correlated with engagement for podcasts with low absolute numbers of streams than for the most popular podcasts, suggesting that listeners may be less sensitive to style in podcasts made by well-known creators.
 We also identify those linguistic factors that correlate with our engagement metric across the popularity spectrum, and those that are limited to podcasts within a certain popularity range. 
 
Our predictive models prove that stylistic factors alone play a significant role in determining if a podcast has high or low engagement, achieving an accuracy of $72\%$ in distinguishing between very high engagement (top $25\%$ of podcasts by stream rate in the corpus) and very low engagement (bottom $25\%$) examples. We also show that the overall textual information in podcasts is highly predictive of engagement in this experiment, with an accuracy as high as $81\%$. To understand how style in podcasts compares to other spoken media, we apply our analysis to a corpus of TED talks. Finally, 
we manually examine the highest engagement podcasts in our dataset to  characterize their content.
 
\section{Related Work}

\paragraph{Content-Based Podcast Recommendations}

\newcite{yang2019more} model transcripts with a topic model, and the audio with a representation they trained to predict the non-textual attributes of seriousness and energy. They find that combining these representations improves over the purely topic based model on popularity prediction. This work indicates that stylistic attributes are important factors, and raises the question of whether stylistic features derived from the text are valuable as well. \newcite{tsagkias2010predicting}
develop a framework containing a set of attributes, and compare the proportions of these attributes relative to engagement on iTunes.  Our work follows a similar spirit, but we address some limitations of their study, namely, they use a small set of podcasts ($250$), and manually annotate the attributes for every podcast rather than deriving them from the raw data. Since we derive all features automatically, we limit ourselves to concrete, easily quantifiable features, whereas the above paper considers higher level attributes like `one topic per episode' or `fluent'.

\paragraph{Predicting Performance from Language}

Previous research in natural language processing has explored the connections between textual features and audience engagement in books \cite{ganjigunte-ashok-etal-2013-success,maharjan-etal-2018-letting}, YouTube \cite{kleinberg-etal-2018-identifying}, news \cite{naseri2019analyzing}, TED talks \cite{tanveer2018awe}, and tweets \cite{tan-etal-2014-effect,lampos-etal-2014-predicting}. Other works have modeled the relationship between text and various performance metrics such as movie quote memorability \cite{danescu-niculescu-mizil-etal-2012-hello},  forecasting ability \cite{zong-etal-2020-measuring},  congressional bill survival \cite{yano-etal-2012-textual},
success of job interviews \cite{naim2016automated},
and impact of academic papers \cite{yogatama-etal-2011-predicting,li-etal-2019-neural}, in addition to the entire field of sentiment and opinion mining of  data such as user reviews \cite{pang-etal-2002-thumbs}.

\section{Dataset Construction}
\label{sec:dataset}

The Spotify Podcast Dataset \cite{clifton-etal-2020-100000,trec2020podcastnotebook} is a recently released corpus of over $100,000$ podcast episodes, mostly in English, that are transcribed with Google's Speech to Text commercial speech recognition, reported in the paper to have an $18\%$ word error on podcasts.
A podcast, also known as a `show' in the dataset, is a collection of episodes.
In addition to the speech transcripts, the textual information associated with each podcast episode includes the title and description of the episode and the parent show (Table \ref{tab:dataschema}). In this paper, we consider descriptions and transcripts as the text representation of an episode.
All textual data was normalized and part-of-speech tagged with spacy.\footnote{\url{spacy.io} \cite{spacy}, with the large English web model, en\_core\_web\_lg v.2.3.1.} 

\begin{table*}[ht!]
\scriptsize
    \centering
    \begin{tabular}{|l|p{5.1in}|}
    \hline
     Show title    & Witch Wednesdays \\
      \hline
     Show description     & A weekly podcast covering all things witchcraft in the modern world. Join us, two best friends and Midwestern witches (one Wiccan, one not), as we dive into all things witchy. We're starting at the beginning, making this podcast a great resource for newbies...\\
     \hline
     Episode title & Episode 1 - What You’re In For This Year \\
      \hline
     Episode description & Happy New Year! Welcome to Witch Wednesdays! Join us every Wednesday morning for all things witch and witchcraft. In this first episode, we’re introducing ourselves and this podcast so you can get an idea about what you’re getting yourself into this year...\\
      \hline
     Automatic transcript & You're listening to which Wednesday's your weekly podcast source for all things witchcraft in the modern world. Join your host Stephen Tara every Wednesday morning that they dive into a new Wiki topic. Hello and welcome to the very first episode of which Wednesdays. I'm Steph and I'm Terrell and together will be co-hosting this podcast Adventure this year rather than...\\
     \hline
    \end{tabular}
    \caption{Example of textual content (truncated) associated with a podcast in the dataset.}
    \label{tab:dataschema}
\end{table*}

\subsection{Ads and promotions} 
\label{sec:ecmodel}
Since many episode descriptions contain promotions, advertisements, and show notes, which are extraneous to the main content of the podcast, we remove such  material before analysis (although we also measure the amount of ad content as a feature).\footnote{Initial experiments showed weaker effects of stylistic features on engagement when such extraneous content was included in the analysis.} Promotional and extraneous material was detected by the classifier described by \newcite{reddy-etal-2021-detecting}, a model using BERT with a classification head, trained on a manually annotated set of episode descriptions. This classifier is reported to have a sentence classification accuracy of $95\%$ on episode descriptions.

\subsection{Engagement metric} 
We obtained streaming numbers for the episodes in the corpus from Spotify, a music and podcast streaming platform. The numbers were aggregated from the date of the episode's publication on the platform until December 2020. Since the most recently published episode in the dataset is from February 2020, all episodes had several months of exposure by the time of collection.

We specifically consider streaming by `first-time listeners' who are not already familiar with the show, i.e., those who have not previously streamed any other episode of that show for more than five minutes. Listeners who are familiar with the show through other episodes are ignored since they may be habituated and primed for the content. 
As described in the introduction, we use stream rate as the engagement metric, defined as the proportion of the show's first-time listeners who stream at least five minutes of the episode.
Stream rate in the dataset shows a weak but statistically significant inverse rank correlation with popularity (Spearman's $\rho = -0.12$, $p<0.001$). This may be because popular podcasts attract more listeners who may realize they are not interested in the content soon after they begin streaming, while the listeners of less popular podcasts may have actively sought them out. $70\%$ stream rate in a well-known podcast which would have attracted a broad array of listeners is not comparable to $70\%$ stream rate in a relatively unknown podcast. Therefore, we bin the dataset into popularity  quartiles for analysis on stream rate, which is found to be uncorrelated with popularity within each quartile. Stream rate is uncorrelated with the time of publication.

\subsection{Filters}
We filter out all episodes that are shorter than ten minutes and fewer than a threshold number of total streams. 
To control for duration effects in the analysis of transcripts, we truncate transcripts at ten minutes. 
The original podcast corpus contains multiple episodes for many of the show while other show have only one episode. We select the most-streamed episode from each show as its representative, thereby ensuring that every show is represented by a {\em single} episode in the data. This is done so that shows with several episodes do not have an outsize influence on the models.

Since the original corpus is an English-language collection, all of our analysis is constrained to English, and we filter out any stray examples in the corpus that are detected as non-English after running language identification \cite{lui-baldwin-2011-cross} on the descriptions. The resulting dataset has $5371$ episodes.

\subsection{Topics and Genre}
The norms of language usage may vary depending on the genre and topics being discussed. 
For example, technical podcasts are expected to contain more complex language compared to chit-chat, crime podcasts to contain words with negative sentiments as opposed to motivational podcasts, and so on. 
The RSS feed of a podcast show contains one or more categories selected by the creators from the Apple iTunes taxonomy; however, these are unreliable, since many of the categories are ambiguous or ill-defined, 
(e.g. `Leisure' which mainly includes gaming podcasts but also general leisure topics, `Kids \& Family' which includes podcasts for kids as well as about parenting), 
and podcast creators may not always select the most appropriate categories \cite{sharpe2020review}. Furthermore, podcasts span multiple themes and structures, making the assignment of one or two categories per podcast too restrictive.

Instead, we fit an LDA topic model \cite{blei2003latent} with $100$ topics\footnote{The number of topics is selected by optimizing for topic coherence as implemented by the coherence model in the Gensim toolkit \cite{rehureklrec}.} to  transcripts of the entire 100k podcast corpus as in previous works \cite{clifton-etal-2020-100000,yang2019more}, 
represent each episode by the topic distribution, and measure topic proportions relative to the target metrics in order to contextualize our results on stylistic features. Table \ref{tab:lda} shows a sample of the inferred topics.

\begin{table}[ht!]
\scriptsize
   \centering
    \begin{tabular}{|l|l|}
\hline
     Genre    & Words in Topic \\
     \hline
     mystery & door, eye, room, hand, head, night, face, away, looked \\
     music & song, music, album, artist, listen, love, record, hip, hop \\
     investing & market, company, stock, investment, investor, trade\\
     working out & training, gym, fitness, coach, workout, muscle, body \\
     entertainment & jacob, alice, edward, vampire, max, bella, hamilton, john  \\
     ad & free, episode, app, download, podcasts, listen, place \\
     culture & world, sort, idea, human, interesting, sense, fact, society \\
     education & school, student, class, teacher, college, high, kid, grade \\
     gaming & game, play, playing, new, nintendo, stuff, played, switch \\
     food & food, eat, coffee, drink, chicken, restaurant, beer, taste\\
     tv & episode, character, show, scene, season, end, point \\
     harry potter & harry, mr, potter, charlie, ron, fred, hermione, professor\\
     career & job, company, team, working, career, industry, experience \\
     sports & world, team, australia, cup, final, club, week, player \\
     biology & cell, dna, bond, virus, genetic \\
     crime & murder, police, crime, case, found, death, killer \\
     language & word, language, english, spanish, use, learn, speak \\
     astronomy & space, science, earth, planet, light, solar, scientist, star \\
     fillers 1 & yeah, oh, okay, yes, exactly, gonna, feel, guess, sure, cool \\
     fillers 2 & feel, stuff, still, never, went, remember, thought, whatever \\
     effusiveness & love, great, thank, different, amazing, bit, awesome \\
     \hline
    \end{tabular}
    \caption{Some examples of LDA topics. The genre labels are manually assigned only to aid interpretation.}
    \label{tab:lda}
\end{table}

\section{Linguistic Features}
\label{sec:features}
We define a set of explainable linguistic features that are hypothesized to affect engagement. These features have been drawn from different podcasting advice blogs, alongside some of our own intuitions. 

\paragraph{Length}

Descriptions are known to be important for listeners on their first encounter with the podcast. We also measure audio duration, since surveys show it is a consideration  \cite{mclean_2020}.

\paragraph{Proportion of ads and show notes}
Descriptions of well-known podcasts tend to contain advertisements of other podcasts made by the same network, links to the hosts' or guests' social media presence and websites, or show notes and transcripts, and podcast creators are often advised to include such information \cite{dennis}, and surveys have shown that the majority of podcast listeners do not mind sponsor ads in the content \cite{mclean_2020}. 
We measure the the proportion of text detected on episode descriptions by the  extraneous content classifier described in \S\ref{sec:ecmodel}. 
The proportion of ads in transcripts is given by a manually identified LDA topic that corresponds to words indicative of ads.

\paragraph{Faithfulness of episode descriptions to transcripts}
Length is a weak signal of informativeness. Do listeners seem to prefer descriptions that accurately convey the topics and synopsis of the episode? 
We measure faithfulness of the episode description to the first ten minutes of the transcript as the cosine similarity between the TF-IDF bag of words representation of both texts. While we do not have ground-truth labels to evaluate this definition of faithfulness, we assessed it to be a good heuristic by anecdotally reviewing some examples.\footnote{We found that BERT and related pretrained transformer models are not well suited for this similarity estimation, possibly because of speech recognition errors in the transcripts. If ground-truth faithfulness labels were available, such models could  be trained to make accurate judgments.}

\paragraph{Distinctiveness}
Podcast creators are often encouraged to develop a distinctive style \cite{gray_2021}.  We define distinctiveness as the perplexity of the given text under a unigram language model trained over all the episodes in the dataset. To control for length, we follow the protocol in \newcite{zhang-etal-2019-finding} of randomly sampling a constant number of words from each text and taking the mean cross entropy over a few samples.\footnote{The text was lightly normalized by case-folding and replacing URLs and social media handles with special tokens. We fixed the constant number of words as $100$ for descriptions and $1000$ for transcripts, and sampled over $5$ runs.}

\paragraph{Reading Grade Level}
Similarly to \newcite{zong-etal-2020-measuring}, we make two measurements: the Flesch-Kincaid grade level \cite{flesch1948new} that measures the number of syllables per word and the number of words per sentence, and the Dale-Chall grade level \cite{chall1948formula} which measures word `difficulty' using a lookup table. While caution must be taken on interpreting reading grade level for transcribed speech, these measures have  been explored for speech in prior work \cite{schumacher2016readability}.

\paragraph{Vocabulary Diversity}
We examine whether  podcast creators of high engagement podcasts use more diverse vocabularies, quantified by the entropy of the unigram words in the text, motivated by advice to avoid word repetition \cite{bellis2017}.

\paragraph{Sentiment and Emotion}
Popular advice often encourages podcast creators to be upbeat and positive \cite{briggman_2020}.
The NRC Emotion Lexicon \cite{Mohammad13} contains positive and negative sentiment assignments, as well as  emotions such as anger, trust, and fear, for $14182$ words.\footnote{We experimented with the method of \newcite{demszky-etal-2019-analyzing} to expand the lexicon for the domain by training GloVe
embeddings on the dataset, and then for each emotion, retrieving the  words that have the highest mean cosine similarity to the words associated with that emotion. However, an examination of the expansions for our dataset showed that they include too many false positives.
} 
We measure the proportion of words associated to each of the emotions and sentiments.
Since a lexicon lookup for sentiment is naturally limited in that it does not account for compositionality and cannot model words and variants that are missing in the lexicon, we also apply a full-sentence classifier, the sentiment model from the Google Natural Language API\footnote{\url{https://cloud.google.com/natural-language}, accessed Dec 2020.}. 
The output of the classifier is a score between $+1$ and $-1$ for each sentence.
We define positive and negative polarities for each text as as the proportion of sentences in the text with highly positive (over $+0.5$) or highly negative (under $-0.5$) scores.

\paragraph{Syntax}
Syntactic features are measured by the relative frequencies of each part-of-speech tag. While previous work of this nature finds strong effects of syntactic patterns from parses \cite{ganjigunte-ashok-etal-2013-success}, we find that the noisy speech transcripts 
result in particularly noisy parses from off-the-shelf parsers.

\paragraph{Swearing and fillers}
We conjecture that podcasts with swearing and adult language may not have broad appeal. 
Public speaking recommendations  in podcasting guides \cite{coips_kramer_2020} emphasize the reduction of filler words like `yeah' or `okay', and the use of professional speech.
We attempted to manually define lexicons of these types of categories, but found that it is challenging and prone to human biases, especially given the novel domain and automatic transcripts. Instead, we take advantage of the observation that some of the topics inferred by the LDA model correspond to 
swear words and filler terms,
and measure the proportions of these topics.

\paragraph{Speech Rate and Non-Speech Time}
Podcast creators are often encouraged to speak slowly, since novice speakers tend to rush their delivery \cite{gray_2021b}. Since the transcripts in the dataset contain time alignments of each word, we measure the duration of speech segments in the audio, giving us the speech rate in terms of words per minute. We also measure the amount of time spent on non-speech.

\section{Models and Analysis}

\subsection{Group Means Differences}
In this section, we analyze the different linguistic features by comparing group means between the top  and bottom 25\% of podcasts by engagement within each popularity quartile (approximately 335 podcasts per group) with bootstrapped Welch's t-tests.  
We report the group mean differences of LDA topic proportions  in order to contextualize results on the other features.
For LDA features, we note significance after a Bonferroni correction of $\alpha=0.05/100$, and for the other linguistic features, a Bonferroni correction of $\alpha=0.05/30$.

In the results, `description' refers to the concatenation of the show description and the representative episode's description. When there is an effect from the show description but not the episode's or vice versa, they are explicitly identified as such.
 
 \subsubsection{Genres}
 \label{sec:genresucess}
 
 Among the podcasts in the top popularity quartile, high engagement is associated with topics around lifestyle and culture, mental health, spirituality, and crime, while in the lower popularity quartiles, high engagement podcasts include those about investing, working out, careers, business, parenting, health, art, and relationships.  
 
 \subsubsection{Linguistic Features}
 
\begin{table}[ht!]
\scriptsize
    \centering
    \begin{tabular}{|l|l|l|l|l|}
    \hline
      Measurement   &   \multicolumn{4}{|c|}{Popularity quartile} \\
         \cline{2-5} 
         & 1 (top) & 2 & 3 & 4  \\
         \hline
 \multicolumn{5}{|l|}{\bf Length and duration} \\
 \hline
 Audio duration &  $\uparrow$ &
 $\uparrow$ &
 $\uparrow$ &
  $\uparrow$\\
   Non-speech time in first 10 min &  $\downarrow$ &
   $\downarrow$ & $\downarrow$ & $\downarrow$  \\
        Length of descriptions &  $\uparrow$ & $\uparrow$ & $\uparrow$ &        $\uparrow$ \\
       \hline
 \multicolumn{5}{|l|}{\bf Proportion of ads} \\
  \hline
 Episode description &  $\uparrow$ & & & \\
 Transcript & $\downarrow$ & $\downarrow$ & $\downarrow$ & \\
 \hline
{\bf Faithfulness of description} 
   &  $\uparrow$ &$\uparrow$  & $\uparrow$ & $\uparrow$  \\
 \hline

               \multicolumn{5}{|l|}{\bf Distinctiveness} \\
               \hline
    Descriptions &  $\downarrow$ & $\downarrow$ & $\downarrow$ &  $\downarrow$ \\
      Transcript &  $\downarrow$ & $\downarrow$ & $\downarrow$ &        $\downarrow$ \\
       \hline
         \multicolumn{5}{|l|}{\bf Reading grade level} \\
         \hline
         Descriptions: Flesch-Kincaid &  $\uparrow$ & $\uparrow$ & $\uparrow$ & $\uparrow$ \\
          Descriptions: Dale-Chall &  $\uparrow$ & $\uparrow$ & $\uparrow$ & $\uparrow$ \\
        Transcript: Flesch-Kincaid  &  & $\uparrow$ & $\uparrow$ & $\uparrow$  \\
          Transcript: Dale-Chall &  & $\uparrow$ & $\uparrow$ & $\uparrow$  \\
         \hline
  \multicolumn{5}{|l|}{\bf Vocabulary diversity} \\
  \hline
  Descriptions & $\uparrow$ & $\uparrow$ & $\uparrow$ & $\uparrow$\\
  Transcript & $\uparrow$ & $\uparrow$ & $\uparrow$ & $\uparrow$\\
  \hline
       \multicolumn{5}{|l|}{\bf Word-level sentiment and emotion} \\
    \hline
    Positive sentiment in transcript &  &  &  & $\uparrow$ \\
    Trust in descriptions &  & $\uparrow$ & & $\uparrow$\\
    Trust in transcript&  & $\uparrow$ & $\uparrow$ & $\uparrow$ \\
    Joy in transcript& &  & & $\uparrow$ \\
    \hline
    Anticipation in descriptions &  $\uparrow$ & & $\uparrow$ &  $\uparrow$\\
    Anticipation in transcript &   & &  & $\uparrow$ \\
    Surprise in transcript &  & $\downarrow$ & $\downarrow$ & \\
    \hline
    Negative sentiment in descriptions &  $\uparrow$ & $\downarrow$ & & \\
    Negative sentiment in transcript &   & $\downarrow$ & $\downarrow$ & $\downarrow$  \\
    Fear in descriptions &  $\uparrow$& $\uparrow$& & $\uparrow$ \\
    Fear in transcript &  & $\downarrow$ & $\downarrow$ &$\downarrow$ \\
    Sadness in transcript &  & $\downarrow$ & $\downarrow$ & \\
    Anger in transcript&  $\downarrow$ & $\downarrow$ & $\downarrow$ &$\downarrow$\\
    Disgust in transcript&   & $\downarrow$ & $\downarrow$ & $\downarrow$ \\
    \hline
     \multicolumn{5}{|l|}{\bf Sentence-level sentiment}\\
    \hline
    Positive in descriptions&  & $\uparrow$ & &  \\
    Positive in transcript&  & $\uparrow$ & $\uparrow$&$\uparrow$  \\
    Negative in descriptions&  & $\downarrow$ &$\downarrow$ & $\downarrow$\\
    Negative in transcript&  & $\downarrow$ &$\downarrow$ & $\downarrow$\\
        \hline
       \multicolumn{5}{|l|}{\bf Syntax} \\
    \hline
    Adjectives in descriptions & $\downarrow$ & $\downarrow$ & $\downarrow$ & $\downarrow$ \\
     Adpositions in transcript & $ \uparrow$ &$\uparrow$ &$\uparrow$ &$\uparrow$ \\
     Adverbs in descriptions & $\downarrow$ &  & $\downarrow$ & $\downarrow$ \\
     Adverbs in transcript & $ \uparrow$ &$\uparrow$ &$\uparrow$ &$\uparrow$ \\
Conjunctions in transcript & $ \uparrow$ &$\uparrow$ &$\uparrow$ &$\uparrow$ \\
    Determiners in transcript&  $ \uparrow$ &$\uparrow$ &$\uparrow$ &$\uparrow$ \\
    Interjections in transcript&  $ \downarrow$ & $\downarrow$ & $\downarrow$ & $\downarrow$ \\
          Nouns in descriptions  & $\downarrow$ &  & $\downarrow$ & $\downarrow$ \\
        Nouns in transcript & & & $ \uparrow$ &$ \uparrow$\\
Pronouns in descriptions & $\downarrow$ &  $\downarrow$ & & \\
      Pronouns in transcript &  & & & $ \downarrow$ \\
      Particles in transcript&  & $ \uparrow$ & $ \uparrow$ & \\
    Proper nouns in transcript&  $\downarrow$ & $\downarrow$ & $\downarrow$ & $\uparrow$ \\
  Punctuation in descriptions & $\uparrow$ & $\uparrow$ & $\uparrow$ & $\uparrow$ \\
    \hline
       \multicolumn{5}{|l|}{\bf Swearing and fillers in transcripts} \\
    \hline
    Swearing &  $\downarrow$ & $\downarrow$ & $\downarrow$ & $\downarrow$\\ 
    Fillers &  & &  & $\downarrow$\\ 
    \hline
          {\bf Speech rate} &  $\uparrow$ &
    $\uparrow$ & $\uparrow$ &$\uparrow$ \\
          
    \hline
   
         \end{tabular}
    \caption{Group mean differences between linguistic features of high and low engagement podcasts in each popularity quartile, with the $\uparrow$ ($\downarrow$) arrow indicating increase (decrease) in mean value of the feature for the high group compared to the low. Differences that are not significant after a Bonferroni correction ($p < 0.05/30$ for linguistic features, $p < 0.05/100$ for LDA) are left blank.}
    \label{tab:groupmeans}
\end{table}

Table \ref{tab:groupmeans} shows the features with significant differences across between the high and low engagement groups. We review the main takeaways from these results.

\paragraph{High engagement podcasts are longer, and have appropriate descriptions}
Across all quartiles, podcasts with high engagement tend to be longer on the whole (contrary to advice to keep episodes short), and contain less non-speech in the first ten minutes than the low engagement group. 
They also have descriptions that are more similar to the first ten minutes of the transcripts, which may be because long, faithful descriptions better prepare listeners for the episode.

\paragraph{The correlation between ads and engagement is mixed}
Large amounts of ads in transcripts are associated with lower engagement in all but the bottom popularity quartile. While this may be explained by the fact that many listeners  skip over ads in the audio stream \cite{reddy-etal-2021-detecting}, the effect is strong enough to indicate that ads seem to hurt engagement, even though surveys report that most listeners do not mind ads. The negative association could be a result of our dataset being constrained to first-time listeners; further analysis needs to be done to understand if it holds of returning listeners.
Ads in episode descriptions, on the other hand, do not hurt engagement on the whole, and in fact, are associated with higher engagement in the top quartile, likely because much of the detected `ad' content in popular podcasts consists of promotional material about the podcast itself, which often includes useful information such as links to the hosts' websites and show notes. 

\paragraph{High engagement podcasts tend to use diverse and mainstream language}
 Vocabulary diversity in descriptions and transcripts  is consistently larger in the high engagement group, as is reading grade level. High engagement podcasts have more punctuation in their descriptions and more conjunctions (arising from the use of long sentences), adverbs, adpositions, and determiners in their transcripts. These syntactic features correlate with the topics such as culture, mental health, investing, and art.
 
 At the same time, surprisingly, high engagement podcasts use less distinctive language compared to the rest of the corpus than the low engagement group. On closer examination, we find that podcasts scoring low on reading grade level also score high on distinctiveness. 

\paragraph{High engagement podcasts tend to contain positive sentiments and suspense}
On the whole, high engagement is associated with more positive and less negative emotions and sentiment. This relationship is stronger outside of the top popularity quartile.
A notable exception is `fear' in the top popularity quartile, which is explained by the high engagement of popular crime-related podcasts. 

\paragraph{High engagement podcasts are less likely to contain interjections and swearing}
As expected, words such as `oh', `right', and `cool' in contexts that the tagger infers as interjections are significantly less likely to occur in high engagement podcasts. 
Similarly, swearing is associated with low engagement.
Filler words are only negatively associated with engagement in the lowest popularity quartile, though the lack of correlation in other quartiles could be because the LDA topics representing fillers don't model context, and therefore do not capture their discourse function in the way the tagger does for interjections.

\paragraph{High engagement podcast creators tend to speak relatively fast}
While popular advice warns presenters against rushing their speech, the data indicates that on average, high engagement is associated with high speech rates, which is also a finding in previous work  \cite{tsagkias2010predicting}.

\subsection{Predictive Models}
Next, we build classifiers to automatically distinguish high and low engagement podcasts. The prediction task is treated as a balanced binary classification problem.  We make a single dataset for podcasts across all quartiles by aggregating the top and bottom $K\%$ podcasts by stream rate within each quartile. This aggregation is to ensure fair comparisons of podcasts in different quartiles, since a stream rate value that is considered high for a popular podcast, for example, may not be so in the low quartiles. Models are trained and evaluated with the same stratified 5-fold cross validation splits.

We train logistic regression classifiers using different representations of the content: the linguistic features listed previously, the non-stylistic LDA topic proportions, and bag-of-ngrams (unigram and bigram words) with TF-IDF scoring. 
In addition, we train two neural classifiers -- a feedforward neural network with a single hidden layer, using a paragraph vector representation \cite{le2014distributed} of the document as input\footnote{Paragraph vector embeddings were trained on the descriptions and transcripts of the full 100k+ podcast corpus}, and the pre-trained BERT \cite{devlin-etal-2019-bert} uncased English model\footnote{We used the implementation in the Hugging Face library \cite{wolf-etal-2020-transformers}, \url{https://huggingface.co/bert-base-uncased}.}  with a classification head, fine-tuned on this task.
With the linguistic features, we also conduct an ablation study,  removing one group of features at a time, to estimate their contributions to predictive performance.
Prediction accuracies (Table \ref{tab:basicprediction}) are over $70\%$ with linguistic features only, indicating that the features that we have identified are relatively strong predictors of engagement. The reading grade level of descriptions and transcripts makes a big contribution as shown in the ablation results, as do the syntactic features on transcripts. 

\begin{table}[ht!]
\scriptsize
    \centering
    \begin{tabular}{|l|l|l|}
    \hline
  \multicolumn{2}{|l|}{\em Features}   &  {\em Accuracy}  \\
  \hline
  \multicolumn{2}{|l|}{Chance} &  50.00 \\
      \hline

 & Descriptions  &  66.52\\
Logistic Regression  & - Reading grade level &   64.51 \\
with Linguistic & Transcript  &  69.24\\
Features & - Reading grade level &   64.99 \\
 & - Part-of-speech & 65.21\\
& Descriptions + Transcript & 71.51\\
\hline
  \multicolumn{2}{|l|}{Logistic Regression with Non Stylistic LDA Topics} &  72.41 \\
 \hline
Logistic Regression  & Descriptions &  75.35\\
with Bag-of-Ngrams & Transcript&  75.98\\
 & Descriptions + Transcript & 76.25 \\
 \hline
Feed-Forward Network & Descriptions &  74.31\\
with Paragraph Vectors  & Transcript  &76.03 \\
 & Descriptions + Transcript & 77.85 \\
 \hline
 & Descriptions &77.33  \\
BERT & Transcript & 78.52\\
& Descriptions + Transcript & 80.54 \\
\hline
    \end{tabular}
    \caption{Accuracy of predicting whether a podcast is high or low engagement (top or bottom $25\%$ by stream rate), averaged over 5 cross validation splits. Ablation results, shown by `- feature group', are included when there is a significant difference of more than 1 percentage point from the full reference feature set. All pairwise differences are significant.}
    \label{tab:basicprediction}
\end{table}

Analysis of the weights of the bag of n-grams models surface patterns in language usage that corroborate our analysis on linguistic features --  swearing and  negative sentiment is predictive of low engagement, for example. They also suggest  subtle dimensions of variation to complement our set of linguistic features. In Table \ref{tab:bow}, we collect some of the most predictive terms and manually group them into classes. First or second person pronouns are predictive of high engagement in contrast to third person pronouns. This aligns with the finding by \newcite{tsagkias2010predicting} that personal experiences are favored in high engagement podcasts. While fillers exist in both groups, the specific terms used are different, with `kind of' and `literally' being predictive of high engagement in contrast to `um' and `but like'. 
The conjunction `and' is preferred by high engagement podcasts over `but', and `so' over `because'. Interrogative words are more predictive of high engagement with the exception of `which', as are open-ended and future looking terms like `asking', `explore', and `started' over grounded, immediate terms like `make', `use', `today', and `quickly'. 
We emphasize that this is a small qualitative analysis of the most predictive features, and more work needs to done to establish which terms are actually used in semantically similar contexts in the data. We leave explorations of computable features that encode these aspects to future work. 

\begin{table}[ht!]
\scriptsize
    \centering
    \begin{tabular}{|p{1.4in}|p{1.4in}|}
    \hline
    {\em Low engagement} & {\em High engagement} \\
    \hline
    he, she, they, his, her, him,  it & me, you, us, we, my, our, their, myself, someone \\
        \hline
        um, gonna, oh,  like like, because like, but like, such as, okay, all right, you guys, basically
& 
and and, sort of, kind of,  was like, you know, quite, literally
\\
\hline
 but, because & and, so \\
\hline
which & when, what, who, how \\
 \hline
 all & lot of,  little bit \\
 \hline
 says & asking \\
 \hline
 can, cannot & was, were, wasn't \\
 \hline
 make, use  & explore, wanted \\
 \hline
 today, still, quickly & always, started, the time \\

         \hline
    \end{tabular}
    \caption{Terms in descriptions and transcripts sampled from the top 200 unigrams and bigrams that are highly predictive of engagement. The terms are manually arranged to indicate contrasting usage of similar classes of words for qualitative analysis.}
    \label{tab:bow}
\end{table}

On the whole, models with lexical content features perform better than the linguistic signals, which is expected since these models encode more information than a small set of hand-designed features. The BERT classifiers achieve nearly $81\%$ accuracy, indicating that podcast content is highly predictive of engagement. 

Table \ref{tab:K} shows how classification accuracies change when the task is to distinguish the top and bottom $K\%$ podcasts, with $K$ ranging from $10$ to $50$ (all reports thus far have been with $K=25$). Performance drops as $K$ increases (and the gap between the two sets thereby decreases) although the amount of training data goes up, showing that the differences in language usage are more predictable at the extremes of engagement.

\begin{table}[ht!]
\scriptsize
    \centering
    \begin{tabular}{l|c|c|c|c|c|}
         & K=10 & K=15& K=20&K=25 & K=50  \\
         \hline
All Linguistic Features & 74.72 &73.66 & 71.85&71.51 & 63.33  \\
Bag-of-Ngrams & 79.66 & 79.07 &78.15& 76.25 & 69.03 \\
BERT & 83.21 & 83.98 & 81.36 & 80.54 & 68.19\\
\hline
    \end{tabular}
    \caption{Classification accuracy (using descriptions+transcript) tends to goes down as the gap between high and low engagement groups decreases.}
    \label{tab:K}
\end{table}

\section{Podcasting vs Public Speaking: Modeling Engagement with TED Talks}

\begin{table}[ht!]
\scriptsize
    \centering
    \begin{tabular}{|l|l|l|l|l|}
    \hline
    Measurement & \multicolumn{4}{|c|}{Popularity quartile}\\
    \cline{2-5} & 1 (top) & 2 & 3 & 4  \\
    \hline
    \multicolumn{5}{|l|}{\bf Length and duration} \\
    \hline
    Audio duration & $\uparrow$ &  & $\uparrow$ & \\ 
    Length of description &  $\color{red}{\downarrow}$ & $\color{red}{\downarrow}$ & $\color{red}{\downarrow}$ & $\color{red}{\downarrow}$ \\
    \hline
    {\bf Faithfulness of description} &  $\downarrow$ & $\downarrow$  & & $\downarrow$  \\
    \hline
    \multicolumn{5}{|l|}{\bf Distinctiveness} \\
    \hline
    Description &  & $\color{red}{\uparrow}$ & $\color{red}{\uparrow}$ &  $\color{red}{\uparrow}$ \\
    Transcript & $\downarrow$ & $\downarrow$ & $\downarrow$ &  $\downarrow$ \\
    \hline
    \multicolumn{5}{|l|}{\bf Reading grade level} \\
    \hline
    Transcript: Flesch-Kincaid  & & $\color{red}{\downarrow}$ & & \\          
    Transcript: Dale-Chall & & $\color{red}{\downarrow}$ & $\color{red}{\downarrow}$ &  \\
    \hline
    \multicolumn{5}{|l|}{\bf Vocabulary diversity} \\
    \hline
    Description & $\uparrow$ & $\uparrow$ & $\uparrow$ & $\uparrow$\\
    Transcript & $\uparrow$ & $\uparrow$ & $\uparrow$ & $\uparrow$\\
    \hline
    \multicolumn{5}{|l|}{\bf Word-level sentiment and emotion} \\
    \hline
    Positive sentiment in description & $\uparrow$ & & $\uparrow$ & $\uparrow$ \\
    Trust in description & $\uparrow$ & & $\uparrow$ &  \\
    Trust in transcript & & $\uparrow$ & & \\
    Joy in description & & & $\uparrow$ & \\
    \hline
    Anger in transcript & & $\downarrow$ & & \\
    Fear in transcript & & $\downarrow$ & & \\
    Disgust in description & & $\downarrow$ & & \\
    Disgust in transcript & & $\downarrow$ & & \\
    Sadness in transcript & & $\downarrow$ & & \\
    \hline
    \multicolumn{5}{|l|}{\bf Syntax} \\
    \hline
    Adjectives in description& & & $\color{red}{\uparrow}$ & $\color{red}{\uparrow}$ \\
        Conjunctions in transcript &  $\uparrow$ & & & \\
    Particles in description & $\uparrow$ & & $\downarrow$ &  $\downarrow$ \\
       Particles in transcript & $\uparrow$ & & $\uparrow$ & \\
     Pronouns in description& & & $\downarrow$ & \\
    Pronouns in transcript & & & & $\downarrow$ \\
    Punctuation in description& & $\color{red}{\downarrow}$ & $\uparrow$ & \\
\hline
    \end{tabular}
    
    \caption{Significance of group mean differences between linguistic features of higher and lower engagement (top and bottom 25\%) TED talks as given by the proportion of views that left ratings. Red arrows show where the direction of correlation differs from podcasts.}
    \label{tab:tedgroupmeans}
\end{table}

To understand how the relationship between linguistic features and engagement in podcasts compares to other spoken media, we carry out the same analysis on a corpus of $2480$ talks from the TED Conferences \cite{tanveer2018awe,acharyya2020fairyted}. While we don't have access to the stream rate of the lectures, the data includes the total view count and ratings. We define engagement as the proportion of total views that left a rating, with the rationale that the act of leaving a rating is roughly analogous to the podcast engagement metric of listening for several minutes. Another  point of difference between this dataset and the podcasts is that the TED lectures are manually transcribed. Therefore, the data is not directly comparable to the podcast dataset, but we carry out the experiment to try to identify which features of high-engagement speech may be universal, and which are podcast-specific.

We test the same features that we formulated for podcasts, except for LDA topic distributions (due to the small size of the TED corpus relative to the full 100k+ podcast data), and ads and swear words since these occur rarely if at all in TED talks.

Table \ref{tab:tedgroupmeans} shows the group means differences between high and low engagement lectures. On the whole, there are fewer significant differences, because either the TED data is more homogenous than podcasts, the metric isn't directly indicative of engagement, or the features that we designed for podcasts don't apply as much for TED talks.

Like podcasts, higher engagement lectures are longer; however, longer and more faithful descriptions are actually associated with lower engagement. 
Vocabulary diversity is associated with high engagement, but unlike podcasts, high engagement lectures have lower reading grade levels. Since we find that lecture transcripts measure over one grade level higher than podcasts, it could be that after a point, simplicity is rewarded. 
Positive emotions are more significantly associated with engagement compared to the podcast data, which may be because of the inspirational nature of the talks and the relative paucity of crime-related content (and in fact, positive sentiment overall is more prevalent
compared to the podcast data). There is less variation in syntactic features, possibly because  talks are scripted and follow similar templates. The syntactic features with correlations tend to follow similar patterns as in podcasts.


On the prediction task, we achieve up to $71.15\%$ (Table \ref{tab:basicpredictionted}) accuracy using only linguistic features, similar to the performance on podcasts. However, the bag-of-ngrams features are {\em less} predictive than linguistic features, and the BERT model only matches the classifier with linguistic features rather than exceeding it. This may be because there isn't as much variation in topical content as in podcasts.

\begin{table}[ht!]
\scriptsize
    \centering
    \begin{tabular}{|l|l|l|}
    \hline
    \multicolumn{2}{|l|}{\em Features}   &  {\em Accuracy}  \\
    \hline
    \multicolumn{2}{|l|}{Chance} &  50.00 \\
    \hline
    Logistic Regression with  & Description  &  64.01\\ 
    Linguistic Features    & Transcript  &  67.99\\
    & Description + Transcript & 71.15\\
    \hline
    Logistic Regression with    & Description &  67.02\\
    Bag-of-Ngrams   & Transcript &  67.34\\
    & Descriptions + Transcript & 68.40\\
    \hline
    & Descriptions & 68.67 \\
    BERT & Transcript & 66.72\\
    & Description + Transcript & 71.92 \\
    \hline
    \end{tabular}
    \caption{Accuracy of predicting whether a TED talk is high or low engagement.}
    \label{tab:basicpredictionted}
\end{table}

\section{What does Engagement Favor?}
\label{sec:engagement}
Our paper centers five minute stream rate as the target metric for analysis and prediction. Systems optimized for engagement on social media platforms have the potential to spread misinformation and radical content \cite{ribeiro2020auditing}, or be manipulated by bad actors \cite{sehgal2021mutual}. On the other side of the coin, studies have found that algorithms driven by engagement do not spread false news at a higher rate than true news \cite{vosoughi2018spread}, and that 
under certain conditions, engagement metrics may actually reward quality content \cite{ciampaglia2018algorithmic}.

Aggregate stream rate in podcasts is a specific engagement metric distinct from metrics and media in previous studies.
There is limited previous work on engagement in podcasts.  \newcite{holtz2020engagement} find that algorithms driven by engagement lead to less diverse recommendations; however, that work does not study the relationship between the type of content that is favored by the engagement metric. 

While a comprehensive analysis of podcast engagement is beyond the scope of this work, we
manually examine the top $10\%$ of podcast episodes by engagement in our collection, a total of $537$ episodes. As we noted in \S\ref{sec:genresucess}, the LDA topics associated with high engagement are broad: lifestyle, mental health, spirituality, crime, investing, working out, careers, business, parenting, health, art, and relationships. Our manual audit confirms that high engagement podcast do primarily span these topics. In particular, we do not find any episodes containing harmful content, incendiary language, or politically controversial topics in this set. We conclude that while the connection between any absolute measure of intrinsic quality and engagement is unknown, high engagement in our study does not correspond to harmful content.

\section{Conclusion}
This paper presents the first quantitative analysis of how linguistic style and textual attributes in podcasts relate to listener engagement using automatically computed features. We test several hypotheses, and identify  factors that validate popular advice on podcast creation, as well as those with unexpected correlations. Our predictive models perform well at distinguishing high and low engagement podcasts using only textual information. Our comparison with a similar task on TED data shows similarities and differences between podcasts and public lectures vis a vis engagement.

Opportunities for future research include the investigation of other podcast creation advice based on paralinguistic features from the podcast audio (such as pitch and intonation), speaker identities and shifts within a conversation, trajectories of linguistic features over the course of the episode, and models using manual transcripts.

\section*{Acknowledgements}
We thank Ann Clifton, Bernd Huber, Jussi Karlgren, Mi Tian, and Zahra Nazari for their input and discussions.

\section*{Impact Statement}

Since our dataset consists of a few thousand podcasts, uses automatically generated transcripts, and only contains podcasts from publishers owned or operated by Spotify \cite{clifton-etal-2020-100000}, care must be taken when generalizing from these results to deploying automatic recommendation systems, or advising podcast creators. 

It is also worth noting that aggregated engagement data may reflect the language preferences of the dominant community, and may be biased against minority cultural and linguistic subcommunities.
While this dataset lacks self-identified labels on demographics and sociolinguistic identities, there are opportunities for future work (in either podcasts or other media) to collect these self-identifications in order to study questions such as disparities in automatic speech recognition performance by race or gender \cite{koenecke2020racial,tatman-2017-gender}, and whether engagement is biased towards certain dialects.


This paper defined a specific metric, namely, the rate of streaming for at least five minutes; results related to this metric may or may not apply to other engagement metrics.
As with all user data, the engagement metric is influenced by the interface and recommendations of the streaming platform from which the data was collected, and may not translate to other platforms, nor reflect an objective notion of listener engagement. We also reiterate (from \S\ref{sec:engagement}) that listener engagement must not be used as a proxy for intrinsic quality or success.

It must also be emphasized that the stylistic associations that were observed to distinguish high and low engagement podcasts in this particular dataset are correlations with no causality established, and therefore must be interpreted with caution.

\bibliography{anthology,custom}
\bibliographystyle{acl_natbib}
\clearpage

\end{document}